\documentclass{article}

\usepackage[preprint]{tackling_climate_workshop_style}
\usepackage{natbib}

\usepackage[utf8]{inputenc} 
\usepackage[T1]{fontenc}    
\usepackage{hyperref}       
\usepackage{url}            
\usepackage{booktabs}       
\usepackage{amsfonts}       
\usepackage{nicefrac}       
\usepackage{microtype}      
\usepackage{graphicx}
\usepackage{gensymb}
\usepackage{subcaption}

\title{Resolution-Agnostic Transformer-based Climate Downscaling}

\usepackage{authblk}

\title{Resolution-Agnostic Transformer-based Climate Downscaling}

\author[1,1]{Declan J. Curran\thanks{\texttt{d.curran@unsw.edu.au}}}
\author[1,1]{Hira Saleem\thanks{\texttt{h.saleem@unsw.edu.au}}}
\author[1,2]{Sanaa Hobeichi\thanks{\texttt{s.hobeichi@unsw.edu.au}}}
\author[1,1]{Flora D. Salim\thanks{\texttt{flora.salim@unsw.edu.au}}}

\affil[1]{School of Computer Science and Engineering, University of New South Wales, Sydney, New South Wales, Australia}
\affil[2]{ARC Centre of Excellence for the Weather of the 21\textsuperscript{st} Century and Climate Change Research Centre, University of New South Wales, Sydney, New South Wales, Australia}

\begin{document}

\maketitle

\begin{abstract}
Understanding future weather changes at regional and local scales is crucial for planning and decision-making, particularly in the context of extreme weather events, as well as for broader applications in agriculture, insurance, and infrastructure development. However, the computational cost of downscaling Global Climate Models (GCMs) to the fine resolutions needed for such applications presents a significant barrier. Drawing on advancements in weather forecasting models, this study introduces a cost-efficient downscaling method using a pretrained Earth Vision Transformer (Earth ViT) model. Initially trained on ERA5 data to downscale from 50 km to 25 km resolution, the model is then tested on the higher resolution BARRA-SY dataset at a 3 km resolution. Remarkably, it performs well without additional training, demonstrating its ability to generalize across different resolutions. This approach holds promise for generating large ensembles of regional climate simulations by downscaling GCMs with varying input resolutions without incurring additional training costs. Ultimately, this method could provide more comprehensive estimates of potential future changes in key climate variables, aiding in effective planning for extreme weather events and climate change adaptation strategies.
\end{abstract}

\section{Introduction}

Global Climate model simulations are usually produced at coarse resolutions due to the significant computational requirements involved in running these models \cite{wcrp_cmip6}. These models typically operate at resolutions ranging from tens to hundreds of kilometers, which are inadequate for capturing fine-scale atmospheric processes, such as convection, that drive extreme weather events \cite{solman2021cordex}. Additionally, finer resolutions are needed for certain meteorological variables, like soil moisture content and precipitation, to aid in extreme weather events planning and prediction \cite{sun2019microwave,neelesh2024downscaling}.

Recent advances in machine learning (ML) have demonstrated the ability to model the underlying dynamics of the atmosphere. Models like Pangu, Graphcast, Gencast, and NeuralGCM have successfully captured fine-scale meteorological patterns and shown promising results in various climate-related tasks \cite{panguweather2022,graphcast2022,gencast2023,neuralgcm2023}. These models leverage deep learning to enhance the spatial and temporal resolution of climate data, providing insights that were previously unattainable with traditional ML methods. However, most of these models are trained and tested on specific datasets and resolutions, limiting their ability to generalise to new input datasets with finer resolutions. 

In assessing the risks associated with climate change, it’s crucial to consider a comprehensive range of possible changes to key climate variables, such as temperature and precipitation, at regional or local scales \cite{neelesh2024downscaling}. This approach helps address uncertainties in climate models, emissions, and the contributions of natural variability. Traditional methods often rely on using Regional Climate Models (RCMs) to downscale outputs from GCMs in a process known as dynamical downscaling. While effective, this approach is highly computationally intensive \cite{neelesh2024downscaling}. For example, the New South Wales and Australian Capital Territory Regional Climate Modelling (NARCliM) NARCLiM2.0 regional simulations required 14 to 18 months to complete projections to 2100, using the best computational resources available in Australia from the National Computational Infrastructure (NCI) \cite{narclim2021}. This effort cost \$14 million and produced an ensemble of 30 simulations.

Given the limitations of downscaling a large ensemble of GCMs with RCMs, there is a need for alternative approaches that can generate extensive ensembles of plausible future climate scenarios with robust uncertainty estimates \cite{bittner2023lstm},\cite{hobeichi2023using}. Machine learning (ML) offers a promising solution to supplement or enhance existing global or regional climate modeling efforts \cite{neelesh2024downscaling}. 

Given that GCMs are available at various resolutions, an ML model that is agnostic to input resolution could enable the downscaling of a large number of GCMs, regardless of their original resolution. Due to its affordability and significantly lower computational cost at the inference stage compared to running an RCM, such an ML model could facilitate the generation of large ensembles, providing a more comprehensive estimate of potential future changes in rainfall and other key climate variables.

To explore this potential, a pretrained ML-based downscaling algorithm, the Earth Vision Transformer (Earth ViT), was tested on climate/weather data at much higher resolutions than those on which it was originally trained. Additionally, we introduce an amendment to the loss function to penalise results that violate the law of conservation of mass, ensuring that the value of a coarse grid accurately reflects the cumulative property of the enclosed region within its boundary. This step is important for making the ML model more physically consistent \cite{benchmark2024}. 

\section{Earth Vision-Transformer} 
The architecture of the Earth ViT is very similar to the weather forecasting model, Pangu-Weather. The number of pressure levels have been reduced from 13 to 3 and the output head was modified slightly to accommodate a higher resolution necessary for multi-resolution downscaling \cite{panguweather2022}. 

\textbf{Loss Function}
We implemented a custom loss function that ensures the model maintains the conservation of mass property, which is crucial in physical simulations. The loss compares the mass calculated from the low-resolution input images with the mass calculated from the super-resolved output images. Combined with the Mean Squared Error loss, it ensures that the predicted image is as close as possible to the ground truth in terms of pixel values and conserves the same total mass as the input image, enforcing physical consistency.

\begin{center}
    MSE Loss = $\frac{1}{N} \sum_{i=1}^N (y_{true,i} - y_{pred,i})^2 $

    Mass Conservation Loss = $|\sum_{i=1}^N y_{pred,i} - \sum_{i=1}^N y_{input,i}| $
\end{center}

 where $\sum_{i=1}^N y_{pred,i}$ is the total mass (sum of pixel values) in the predicted super-resolved image and $\sum_{i=1}^N y_{input,i}$ is the total mass in the low-resolution input image.

\begin{center}
    Total Loss = MSE Loss + Mass Loss
\end{center} 

\section{Experiments and Results}
We trained Earth ViT and a ResNET on daily ERA5 \cite{hersbach2020era5} climate variables for 2× downscaling tasks from 0.5\degree to 0.25\degree (Approx. 55km to 28km). The Earth-ViT was trained on daily data from 2000/2001 and the ResNET was trained on data from 2000 - 2005. Both models were tested on data from 2006. The ResNETs were only trained on surface data; results are only reported across four surface variables  (10-meter u/v wind components, 2-meter temperature, total precipitation) .

These models were then applied to a similar 2× downscaling task on BARRA-SY data  \cite{barra2022} in 2006 from 3km to 1.5km. Note that the model was not finetuned or trained on any data at this fine resolution, but the climate variables remain the same. As a comparison, one version of the Earth-ViT was also trained directly on BARRA-SY.  Details of training hyperparameters is given in Appendix \ref{hyper}. The details of datasets are given in Appendix \ref{data}.

We evaluated the models across all surface variables using three performance metrics: RMSE, PSNR, and SSIM. RMSE measures prediction accuracy, while PSNR (Peak Signal-to-Noise Ratio) assesses the quality of the reconstructed image by comparing the maximum possible signal to the noise. SSIM (Structural Similarity Index) evaluates the structural similarity between    predicted and reference images, focusing on aspects like luminance and contrast. These metrics provide a comprehensive evaluation of the models' performance \cite{inffus2022}.

The Earth ViT model was compared against ResNET and Bilinear Interpolation, which are established benchmarks in super-resolution tasks. This comparison highlights Earth ViT’s effectiveness in improving downscaling performance\cite{inffus2022}. The performance results for downscaling 2-meter temperature are shown in Table \ref{result1} and \ref{result2}. 
\begin{figure}[htbp]
    \centering
    \begin{subfigure}[t]{0.3\textwidth}
        \centering
        \includegraphics[width=\textwidth]{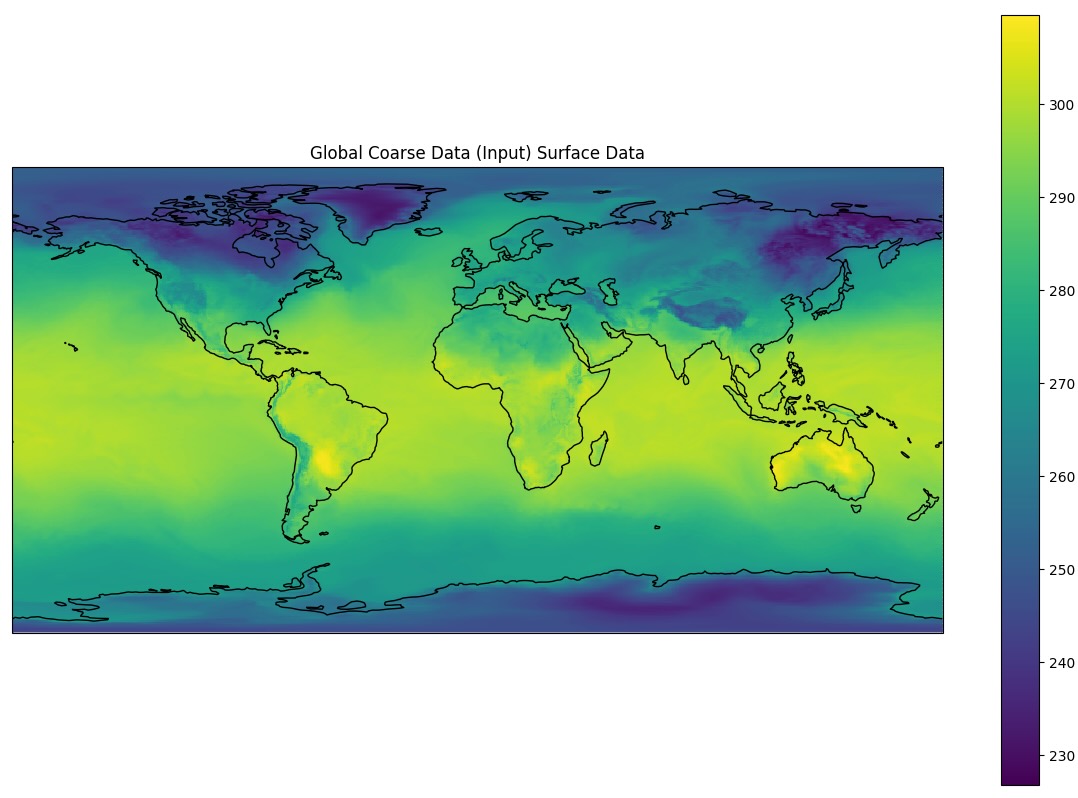}
    \end{subfigure}%
    \begin{subfigure}[t]{0.3\textwidth}
        \centering
        \includegraphics[width=\textwidth]{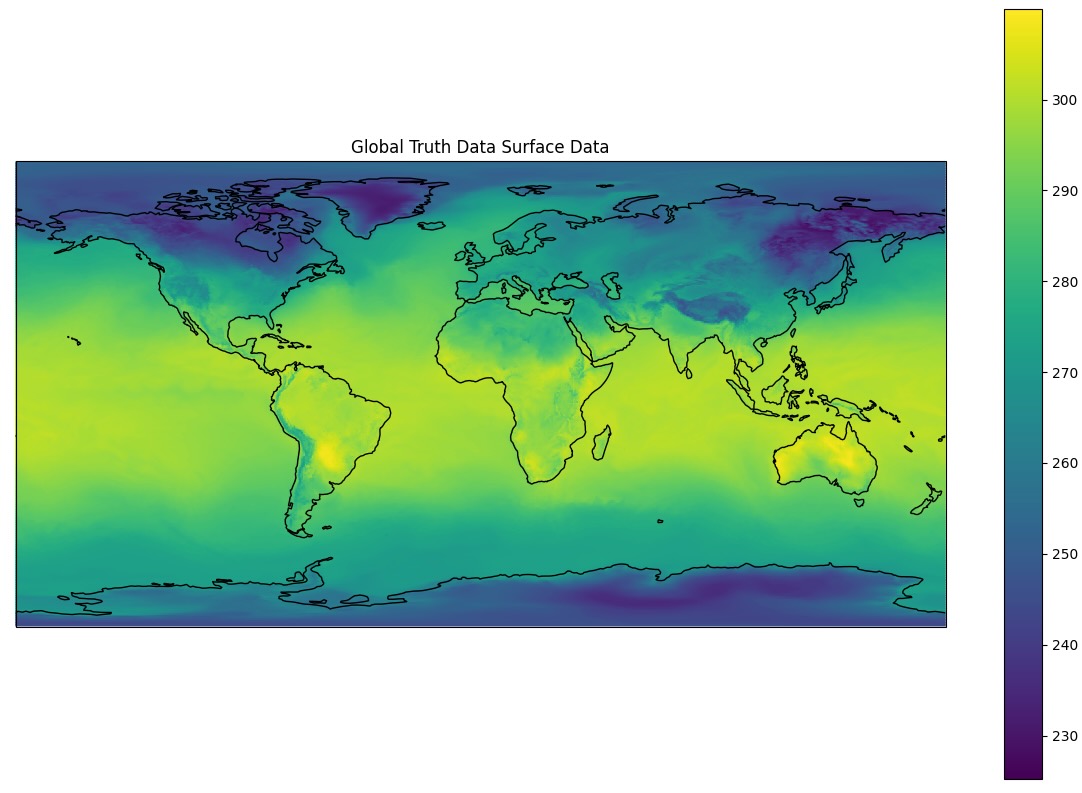}
    \end{subfigure}
    \begin{subfigure}[t]{0.3\textwidth}
        \centering
        \includegraphics[width=\textwidth]{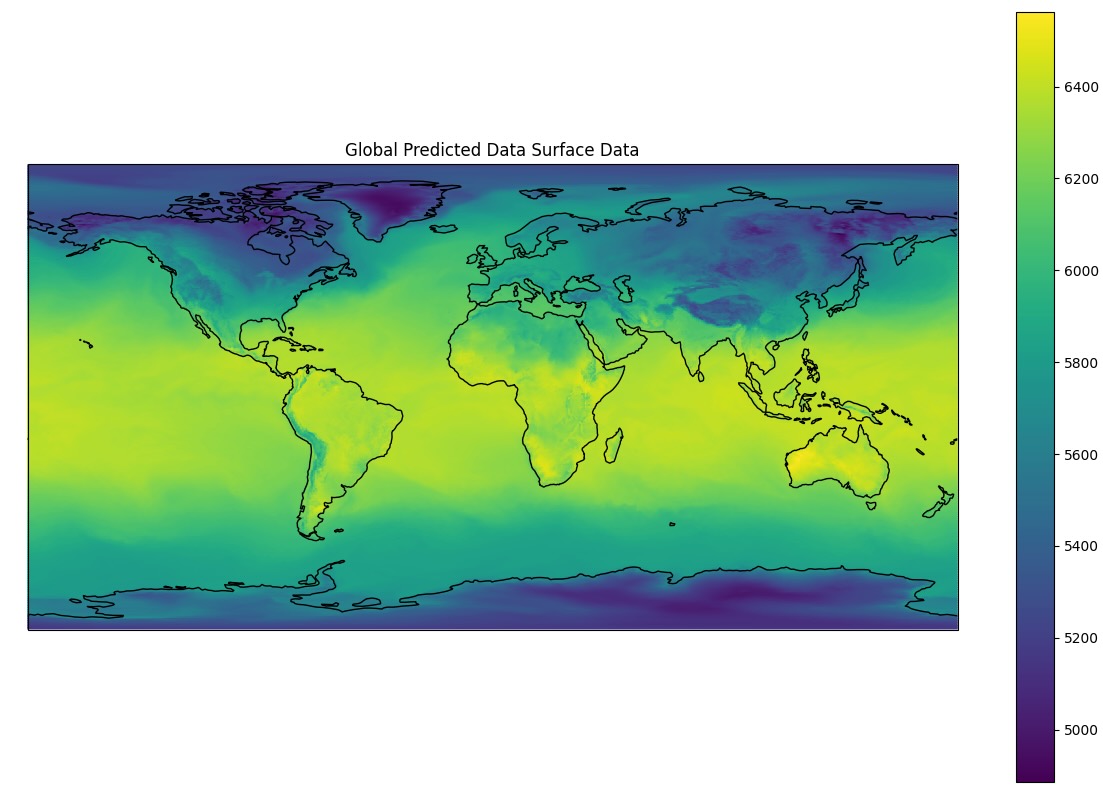}
    \end{subfigure}
    \caption{ERA5 downscaling of 2-meter Temperature using Earth ViT with modified loss function. From left to right: coarse resolution (50km), ground truth high resolution (25km), predicted high resolution (25km)}
    \label{fig:combined}
\end{figure}
\begin{figure}[htbp]
    \centering
    \begin{subfigure}[t]{0.3\textwidth}
        \centering
        \includegraphics[width=\textwidth]{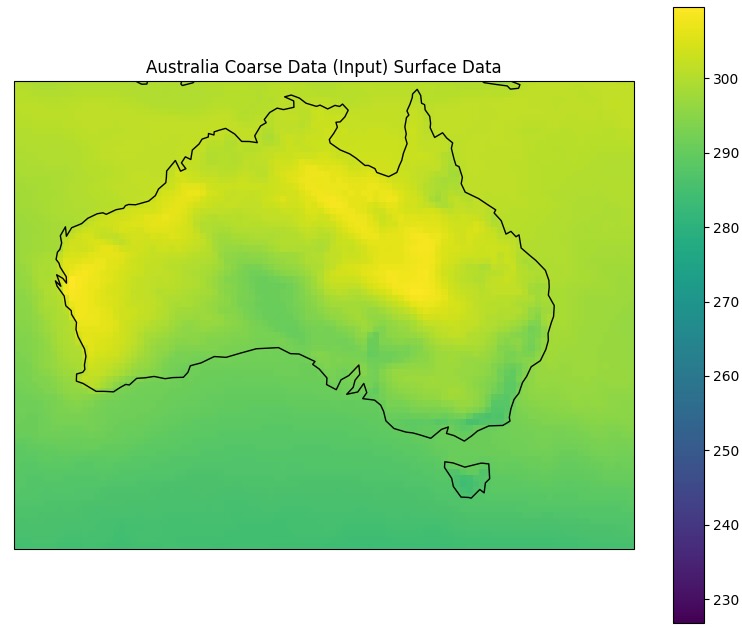}
    \end{subfigure}%
    \begin{subfigure}[t]{0.3\textwidth}
        \centering
        \includegraphics[width=\textwidth]{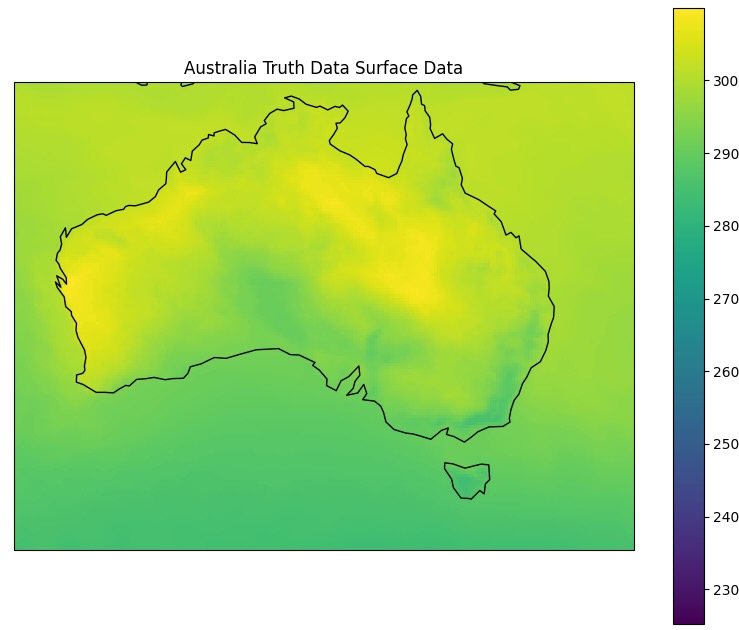}
    \end{subfigure}
    \begin{subfigure}[t]{0.3\textwidth}
        \centering
        \includegraphics[width=\textwidth]{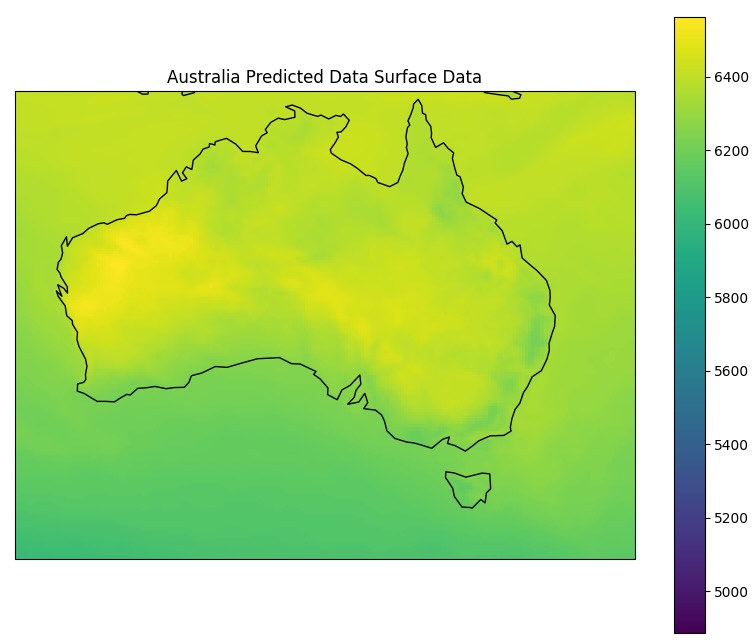}
    \end{subfigure}
    \caption{ERA5 downscaling of 2-meter Temperature using Earth ViT with modified loss function. From left to right: coarse resolution (50km), ground truth high resolution (25km), predicted high resolution (25km)}
    \label{fig:combined}
\end{figure}
Figure1 (Figure2) displays the average 2-meter temperature for the testing year across both the coarse resolution and the fine resolution of the truth and prediction when downscaling ERA5 (BARRA-SY coarsened) from 50km to 25km (from 3km to 1.5km). The results demonstrate a strong overall agreement between the truth and the prediction in terms of both magnitude and spatial consistency.

Results for downscaling precipitation, u\_component\_of\_wind and v\_component\_of\_wind are given in Appendix \ref{results}.
\begin{table}[h]
  \caption{ERA5 Results on Temperature from 50km to 25km downscaling}
  \label{result1}
  \centering
  \begin{tabular}{ |p{5cm}|p{1.5cm}|p{1.5cm}|p{1.5cm}|p{1.5cm}|}
    \toprule
    Model     & RMSE $\downarrow$ & PSNR $\uparrow$ & SSIM $\uparrow$& Carbon $(kg CO_2)$ \\
    \midrule
Bilinear Interpolation & 0.279 & 11.084 & 0.995 & \textbf{0.452} \\ 
\midrule
ResNET & 8.541 & -18.611 & 0.969 & 5.481\\ \midrule
ResNET - Modified Loss & 16.632 & -24.413 & 0.679 & 5.826\\ \midrule
Earth ViT & 0.023 & 32.576 & 0.983 & 1.603\\ \midrule
Earth ViT - Modified Loss & \textbf{0.019} & \textbf{34.061} & \textbf{0.987} & 1.805\\ 
    \bottomrule
  \end{tabular}
\end{table}
\begin{table}[h]
  \caption{BARRA-R Results - Trained on ERA5 - tested on BARRA-R}
  \label{result2}
  \centering
  \begin{tabular}{ |p{5cm}|p{1.5cm}|p{1.5cm}|p{1.5cm}|p{1.5cm}|}
    \toprule
 Model     & RMSE $\downarrow$ & PSNR $\uparrow$ & SSIM $\uparrow$& Carbon $(kg CO_2)$ \\
    \midrule
Bilinear Interpolation & 0.138 & 17.269 & \textbf{0.992} & \textbf{0.002} \\ \midrule
ResNET & 1.082 & 1.647 & 0.984 & 0.002  \\ \midrule
ResNET - Modified Loss & 3.199 & -9.807 & 0.755 & 0.003  \\ \midrule
Earth ViT  & 0.031 & 30.221 & 0.973 & 0.040 \\ \midrule
Earth ViT (TRAINED ON BARRA) & 0.031 & 30.229 & 0.964 & 0.194\\ \midrule
Earth ViT - Modified Loss & 0.031 & 30.375 & 0.983 & 0.043\\ \midrule
Earth ViT - Modified Loss (TRAINED ON BARRA) & \textbf{0.029} & \textbf{30.690} & 0.969 & 0.173\\ 
    \bottomrule
  \end{tabular}
\end{table}

We can see that penalising the model for physical discrepancies make it perform better acorss most of the metrics. In table \ref{result2}, we can see the performance of EarthViT trained on BARRA is almost same as the ERA5 pre-trained EarthViT. Therefore, we can deduce that The pretrained Earth Vision Transformer (Earth ViT) model can generalise well across varying resolutions (50 km and 3 km)  and datasets (ERA5 and BARRA-R); even with less training data. 

\section{Discussion}
 A resolution-agnostic model provides significant benefits for downscaling various generations of climate models, such as CMIP5, CMIP6, and the upcoming CMIP7. Using the same model for each ensemble without needing to alter resolutions or undergo additional training streamlines the process and cuts down on computational costs. By generating large ensembles of regional projections, Earth ViT could enable comprehensive climate risk assessments that are both scalable and cost-effective, benefiting sectors such as agriculture, disaster management, and infrastructure planning

Unlike traditional Regional Climate Models (RCMs), Earth ViT offers a more accessible alternative for producing high-resolution climate projections. The introduction of a mass conservation loss function further improved the model's performance, ensuring adherence to fundamental physical laws. This integration is vital for applying the model to regional climate studies and predicting extreme weather events. Not to mention that each of the models presented here are lightweight with large carbon savings over equivalent dynamical downscaling models. 

Future research will focus on refining Earth ViT to enhance its robustness across different regions and scenarios. The model's potential to adapt to new regions or datasets through transfer learning presents further opportunities for expanding its applicability.

\section{Acknowledgement}
SH acknowledges the support of the Australian Research Council Centre of Excellence for the Weather of the 21\textsuperscript{st} Century (CE230100012). The work was undertaken using resources from the National Computational Infrastructure (NCI), which is supported by the Australian Government.  

\bibliographystyle{ACM-Reference-Format}
\bibliography{ref}

\appendix
\section{Experiments}
\subsection{Hyperparameters}
\label{hyper}
\begin{table}[h]
  \caption{Model Training Hyperparameters}
  \label{hyperparam}
  \centering
  \begin{tabular}{lr}
    \toprule
    Parameter     & Value    \\
    \midrule
    Learning rate & 1e-4     \\
    Scaling factor & 2    \\
    Large kernel size  & 9 \\
    Small kernel size & 3 \\
    No of channels & 64 \\
    No of blocks & 16 \\
    Optimizer & Adam \\
    \bottomrule
  \end{tabular}
\end{table}

\subsection{Hardware Requirements}
All models were trained for 50 Epochs, with the ResNETS on NVIDIA V100 GPUs with 32GB RAM and Earth-ViTs on NVIDIA A100 80 GB RAM GPUs.

\section{Datasets}
\label{data}
\subsection{ERA5}
We used ERA5 reanalysis data \cite{hersbach2020era5}, selecting surface variables (10-meter U/V wind components, 2-meter temperature, total precipitation) and pressure-level variables (geopotential, specific humidity, temperature, 10-meter U/V wind components) at three pressure levels (hPa) (50, 100, and 150). These variables were regridded to 0.5 degrees for downscaling.

\subsection{BARRA-SY}
BARRA-SY is an ultra-high-resolution dataset of the Australian BARRA-R reanalysis dataset \cite{barra2022}. At a 1.5 km resolution, it covers the Sydney and coastal New South Wales (NSW) region (28.0°S to 37.5°S latitude and 147.0°E to 154.0°E longitude). The variables selected were consistent with those used in ERA5. Notably, temperature is recorded at 1.5 meters in BARRA-SY, whereas it is recorded at 2 meters in ERA5. Despite this difference, our results have shown that the model's ability to generalise has not been affected.
\newpage
\section{Results}
\label{results}

\begin{table}[h]
\caption{Results for precipitation (pr), u\_component\_of\_wind (u\_wind), v\_component\_of\_wind (v\_wind)  - Trained on ERA5 }
\label{result-table2}
\centering
\begin{tabular}{p{3cm}|ccc|ccc|ccc}
\toprule
\multicolumn{1}{c}{Model}  &\multicolumn{3}{c}{RMSE $\downarrow$} &\multicolumn{3}{c}{PSNR $\uparrow$}  &\multicolumn{3}{c}{SSIM $\uparrow$}         
\\ \midrule
 & pr & u\_wind & v\_wind & pr & u\_wind & v\_wind & pr & u\_wind & v\_wind
\\ \midrule
Bilinear Interpolation 
& 3.66  & 0.22 & 0.22
& 88.71 & 13.00 & 13.16
& 0.99  & 0.99 & 0.98
\\ \midrule
ResNET 
& 0.001 & 2.05 & 1.71
& 75.17 & -6.26 & -4.65
& 0.97 & 0.92 & 0.92
\\ \midrule
ResNET - Physics loss 
& 0.00 & -9.807 & 0.755
& 72.73 & n/a & n/a
& 0.77 & n/a & n/a
\\ \midrule
Earth ViT  
& 0.12 & 0.034 & 0.039
& 18.22 & 29.25 & 28.15
& 0.99 & 0.98 & 0.98
\\ \midrule
Earth ViT - Physics Loss 
& 0.11 & 0.029 & 0.034
& 18.67 & 30.46 & 29.24
& 0.99 & 0.98 & 0.98
\\ \bottomrule
  \end{tabular}
\end{table}

\begin{table}[h]
  \caption{Results for precipitation (pr), u\_component\_of\_wind (u\_wind), v\_component\_of\_wind (v\_wind) - Trained on ERA5 - tested on BARRA-R}
  \label{result-table2}
  \centering
\begin{tabular}{p{3cm}|ccc|ccc|ccc}
\toprule
\multicolumn{1}{c}{Model}  &\multicolumn{3}{c}{RMSE $\downarrow$} &\multicolumn{3}{c}{PSNR $\uparrow$}  &\multicolumn{3}{c}{SSIM $\uparrow$}         
\\ \midrule
 & pr & u\_wind & v\_wind & pr & u\_wind & v\_wind & pr & u\_wind & v\_wind
\\ \midrule
Bilinear Interpolation 
& 0.07 & 0.14 & 0.14
& 25.72 & 16.97 & 16.92
& n/a & 0.97 & 0.97
\\ \midrule
ResNET 
& 0.45 & 1.43 & 1.88
& 12.2 & -1.70 & -3.46
& 0.88 & 0.93 & 0.92
\\ \midrule
ResNET - Physics Loss
& 0.52 & 2.96 & 3.55
& 7.98 & -9.10 & -10.51
& 0.61 & 0.30 & 0.36
\\ \midrule
Earth ViT  
& 0.05 & 0.04 & 0.04
& 26.90 & 26.82 & 27.38
& 0.90 & 0.97 & 0.97
\\ \midrule
Earth-ViT (TRAINED ON BARRA) 
& 0.06 & 0.03 & 0.03
& 25.95 & 29.61 & 29.85
& 0.90 & 0.97 & 0.96
\\ \midrule
Earth-ViT - Physics Loss 
& 0.02 & 0.04 & 0.04
& 25.17 & 27.21 & 27.30 
& 0.91 & 0.98 & 0.97 
\\ \midrule
Earth ViT - Physics loss (TRAINED ON BARRA) 
& 0.06 & 0.03 & 0.03
& 27.18 & 29.85 & 30.22
& 0.91 & 0.97 & 0.96
\\ \bottomrule
  \end{tabular}
\end{table}

\end{document}